\documentclass{article}
\usepackage{hyperref}
\usepackage{url}
\usepackage[dvips]{graphicx,color}
\usepackage{paralist}
\usepackage{amsmath}
\usepackage{amsfonts}
\usepackage[letterpaper, left=1.25cm, right=1.25cm, top=1.25cm, bottom=1.25cm]{geometry}
\usepackage{subfigure}
\usepackage{multirow}
\usepackage{fancyhdr}
\title{Distance Measures for Reduced Ordering Based Vector Filters}
\author{%
        M. Emre Celebi\\
        Dept. of Computer Science\\Louisiana State Univ., Shreveport, LA, USA\\
        \href{mailto:ecelebi@lsus.edu}{ecelebi@lsus.edu}
       }
\begin{document}
\maketitle
\begin{abstract}

Reduced ordering based vector filters have proved successful in removing long-tailed noise from color images while preserving edges and fine image details. These filters commonly utilize variants of the Minkowski distance to order the color vectors with the aim of distinguishing between noisy and noise-free vectors. In this paper, we review various alternative distance measures and evaluate their performance on a large and diverse set of images using several effectiveness and efficiency criteria. The results demonstrate that there are in fact strong alternatives to the popular Minkowski metrics.

\end{abstract}

\section{Introduction}
\label{sec_intro}

Color images are often contaminated with noise, which is introduced during acquisition (sensor noise) and/or transmission (channel noise). Sensor noise is inherent to all electronic image sensors and is often modeled using Gaussian or Poisson distributions \cite{Smolka06}. Channel noise is caused by various sources including man made phenomena (car ignition systems, industrial machines in the vicinity of the receiver, switching transients in power lines, various unprotected switches, etc.) and natural causes (lightning in the atmosphere, ice cracking in the Antarctic region, etc.) \cite{Plataniotis00}. Channel noise is often characterized by impulsive sequences, which occur in the form of short-duration, high-energy spikes attaining large amplitudes with probability higher than that predicted by the Gaussian density model \cite{Smolka04}. The introduction of such noise into an image not only lowers its perceptual quality, but also makes subsequent tasks such as edge detection and segmentation more difficult. Therefore, noise removal is often an essential preprocessing step in many color image processing applications.
\par
Numerous filters have been proposed for the removal of long-tailed, in particular impulsive, noise from color images \cite{Smolka06,Plataniotis00,Smolka04,Lukac05,Lukac06,Celebi07}. Among these, nonlinear vector filters have proved successful in removing the noise while preserving the edges and fine image details. The early approaches to nonlinear filtering of color images often involved the application of a scalar filter to each color channel independently. However, since separate processing ignores the inherent correlation between the color channels, these methods often introduce color artifacts to which the human visual system is very sensitive. Therefore, vector filtering techniques that treat the color image as a vector field and process color pixels as vectors are more appropriate. 
\par
An important class of nonlinear vector filters is the one based on robust order-statistics. These filters involve the ordering of a set of input vectors inside a predefined sliding window. The lowest ranked input vector in this window is often taken as the output vector. The ordering is commonly achieved using variants of the Minkowski distance. Recently, fuzzy similarity measures \cite{Shen04,Morillas05a,Morillas05b,Morillas07a,Morillas07b,Schulte07a,Schulte07b,Camarena08,Morillas08a,Morillas08b} have been proposed as  alternatives to the more traditional distance measures. Since it is often trivial to convert a similarity measure to a dissimilarity measure, throughout this paper we will use the term 'distance measure' to refer to a function that quantifies the distance or similarity between two vectors.
\par
In this paper, we review various distance measures that can be used to order color vectors in reduced ordering based vector filters. We evaluate the performance of these measures on a large set of images that cover a variety of domains using several effectiveness and efficiency criteria. The rest of the paper is organized as follows. Section \ref{sec_dist} presents an overview of order-statistics based vector filters and distance measures that can be utilized to order color vectors. Section \ref{sec_exp} describes the noise model, filtering performance criteria, and the experimental setup. Finally, Section \ref{sec_conc} gives the conclusions and the future work.

\section{Distance Measures for Reduced Ordering of Color Vectors}
\label{sec_dist}

\begin{figure}[!ht]
\centering
\includegraphics[width=0.14\columnwidth,draft=false]{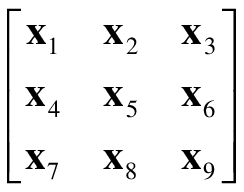}
\caption { \label{fig_window} Indexing convention inside a $3 \times 3$ window }
\end{figure}

Consider an $M \times N$ red-green-blue (RGB) input image {\bf X} that represents a two-dimensional (2D) array of three-component vectors ${\bf x}(r,c) = [x_1(r,c),x_2 (r,c),x_3 (r,c)]$ occupying the spatial location $(r,c)$, with the row and column indices $r = \left\{ 1,  \ldots, M \right\}$ and $c = \left\{ 1, \ldots, N \right\}$, respectively. In the pixel ${\bf x}(r,c)$, the $x_k(r,c)$ values denote the red ($k=1$), green ($k=2$), and blue ($k=3$) components. Most images are nonstationary in nature; therefore filters often operate on the assumption that the local image features can be extracted from a small image region called a sliding window. The size and shape of the window influence the properties and efficiency of the filtering operation and are therefore application dependent \cite{Smolka06}. A square window $W(r,c)$ of size $\sqrt n \times \sqrt n$ pixels centered on ${\bf x}(r,c)$ is commonly used due to its versatility and good  performance \cite{Lukac05}. The window slides over the entire image {\bf X} in a raster fashion and the procedure replaces the input vector ${\bf x}(r,c)$ with the output ${\bf y}(r,c) = F(W(r,c))$ of a filter function $F(\cdot)$ that operates over the samples inside $W(r,c)$. Repeating the procedure for each pair $(r,c)$, with $r = \left\{ 1, \ldots, M \right\}$ and $c = \left\{ 1, \ldots, N \right\}$, produces the output vectors ${\bf y}(r,c)$ of the $M \times N$ filtered image {\bf Y}. For notational simplicity, the input vectors inside $W(r,c)$ are re-indexed as a set, i.e.\ $W(r,c) = \left\{ {\bf x}_i: i=1, \ldots, n \right\}$ (see Figure \ref{fig_window}), as commonly seen in the related literature \cite{Smolka06,Plataniotis00,Smolka04,Lukac05,Lukac06,Celebi07}. In this notation, the center pixel in $W$ is given by ${\bf x}_{(n+1)/2}$ and in the vector ${\bf x}_i = [x_{i1}, x_{i2}, x_{i3}]$ with components $x_{ik}$, the $i \in \{1, \ldots, n\}$ and $k \in \{1,2,3\}$ indices denote the position of the vector inside the window and the color channel, respectively.

\par
Order-statistics based vector filters operate by ordering (ranking) the vectors inside the sliding window. The purpose of this ordering is to distinguish between the noisy and noise-free vectors. However, in contrast to scalar data, there is no universally accepted method for ordering multivariate data. Widely known multivariate ordering methods include \cite{Barnett76}:

\begin{itemize}
	\item Marginal ordering (M-ordering): The vectors are ordered in each component independently. This scheme produces a set of ordered output vectors that is usually not the same as the set of input vectors and consequently results in color artifacts when applied to multichannel image data.
	\item Conditional ordering (C-ordering): The vectors are ordered based on the marginal ordering of one of the components. This scheme disregards the vectorial nature of the multichannel image data.
	\item Partial ordering (P-ordering): The vectors are partitioned into smaller groups that are then ordered. Despite its theoretical appeal, this scheme is computationally demanding \cite{Tang95}.
	\item Reduced (aggregate) ordering (R-ordering): The vectors are first reduced to scalar representatives using a suitable distance  measure. The ordering of these scalars is then taken as the ordering of the corresponding vectors. This is the most common ordering scheme in the literature \cite{Smolka06,Plataniotis00,Smolka04,Lukac05,Lukac06,Celebi07}.
\end{itemize}

To order the vectors inside the sliding window $W(r,c) = \left\{{\bf x}_i: i=1, \ldots, n \right\}$, R-ordering based vector filters first calculate the aggregate (cumulative) distance $D_i$ for each vector ${\bf x}_i$:
\begin{equation}
\label{equ_agg_dist}
D_i = \sum\limits_{j = 1}^n {d({\bf x}_i, {\bf x}_j)}
\end{equation}
where $d({\bf x}_i, {\bf x}_j)$ denotes a suitable distance measure that quantifies the dissimilarity between vectors ${\bf x}_i$ and ${\bf x}_j$. The ordered sequence of scalars $D_{(1)} \leq D_{(2)} \leq \ldots \leq D_{(n)}$ implies the same ordering in the corresponding vectors
${\bf x}_{(1)} \leq {\bf x}_{(2)} \leq \ldots \leq {\bf x}_{(n)}$. Many filters define the lowest ranked input vector (lowest order-statistic), i.e.\ ${\bf x}_{(1)}$, as the output vector. This is because vectors with lower ranks are typical (representative) vectors in their neighborhoods, whereas those with higher ranks are often outliers (noisy vectors) \cite{Pitas91}. More explicitly, the output vector at location $(r,c)$ is given by:
\begin{equation}
\label{equ_reduced}
{\bf y}(r,c) = \mathop {{\rm argmin}}\limits_{{\bf x}_i \in W(r,c)} D_i
\end{equation}
Note that if a similarity measure $s({\bf x}_i, {\bf x}_j)$ is used instead of $d({\bf x}_i, {\bf x}_j)$ in \eqref{equ_agg_dist}, the $\rm argmin$ operator should be replaced with $\rm argmax$.
\par
The vector median filter (VMF) \cite{Astola90}, which is the most well-known R-ordering based vector filter, utilizes the Minkowski distance $d_p ({\bf x}_i,{\bf x}_j )\, = \,\left( {\sum\nolimits_{k = 1}^3 {\left| {x_{ik} - x_{jk}} \right|} ^p } \right)^{1/p}$ as its ordering criterion. Three special cases of $d_p$ are of particular interest: $d_1$ (city-block distance), $d_2$ (Euclidean distance), and $d_\infty$ (chessboard distance). Given the general Minkowski form, $d_1$ and $d_2$ can be defined in a straightforward fashion, while $d_\infty$ is defined as $d_{\infty}({\bf x}_i,{\bf x}_j) = \max_{1\le k\le 3}|x_{ik} - x_{jk}|$. Among these three forms, $d_2$ seems to be preferred in the color image filtering literature due to its isotropic (rotation invariant) nature and good performance \cite{Barni00}. The squared form of $d_2$, i.e.\ $d_2^2$, can also be used as a computationally efficient alternative \cite{Barni92}.
\par
Table \ref{tab_dist} lists some of the traditional distance measures that can be used for the reduced ordering of color vectors. Note that the Bray-Curtis, Canberra, Soergel, and Ware-Hedges measures are normalized forms of the city-block distance, whereas the Divergence Coefficient measure is a normalized form of the Euclidean distance. Some of the measures given in Table \ref{tab_dist} have already been used in the design of various R-ordering based vector filters. For example, the Cosine distance was used in the design of the basic vector directional filter (BVDF) \cite{Trahanias93}. The Goude distance was used in the content-based rank filter \cite{Plataniotis97a}. The rationale behind this measure is that the similarity between two vectors ${\bf x}_i$ and ${\bf x}_j$ can be expressed as the ratio of some function of what they share (commonality) to what they comprise (totality). In particular, the numerator and denominator correspond to the vector difference and the vector sum, respectively.
\par
In \cite{Morillas05b}, the authors introduced a fuzzy magnitude similarity measure given by:
\begin{equation}
\label{equ_fuzzy}
s_\text{fms}^K({\bf x}_i ,{\bf x}_j) = \prod\limits_{k = 1}^3 {{\frac{{\min (x_{ik} ,x_{jk} ) + K}}{{\max (x_{ik} ,x_{jk}) + K}}}}
\end{equation}
where $K$ is a parameter of the measure.
\par
The analogous fuzzy directional similarity measure between vectors ${\bf x}_i$ and ${\bf x}_j$ can be obtained by first normalizing the vectors to unit length, i.e.\ $\hat {\bf x}_i  = {{\bf x}_i \mathord{\left/ {\vphantom {{\bf x}_i {\left\| {\bf x}_i \right\|}}} \right. \kern-\nulldelimiterspace} {\left\| {\bf x}_i \right\|}}$ and $\hat {\bf x}_j  = {{\bf x}_j \mathord{\left/ {\vphantom {{\bf x}_j {\left\| {\bf x}_j \right\|}}} \right. \kern-\nulldelimiterspace} {\left\| {\bf x}_j \right\|}}$, and then calculating the fuzzy magnitude similarity between them, i.e.\ $s_\text{fds}^K({\bf x}_i,{\bf x}_j) = s_\text{fms}^K(\hat {\bf x}_i,\hat {\bf x}_j)$ \cite{Morillas07b}. Furthermore, since both fuzzy magnitude similarity and fuzzy directional similarity values fall into the $(0,1]$ interval, they can easily be combined to obtain a fuzzy magnitude-directional similarity measure as follows: $s_\text{fmds}^{K_1,K_2}({\bf x}_i ,{\bf x}_j) = s_\text{fms}^{K_1}({\bf x}_i,{\bf x}_j) \cdot s_\text{fds}^{K_2}({\bf x}_i,{\bf x}_j)$ \cite{Morillas07b}.
\par
In \cite{Morillas07a}, the authors introduced a combined fuzzy similarity measure given by:
\begin{equation}
\label{equ_cfs}
s^{C,t}_\text{cfs}({\bf x}_i, {\bf x}_j) = \frac{C}
{{C + d_2({\bf x}_i, {\bf x}_j)}} \cdot \frac{t}
{{t + \max \left({\left| {r_i - r_j} \right|,\left| {c_i - c_j} \right|} \right)}}
\end{equation}
where $(r_i,c_i)$ and $(r_j,c_j)$ are the spatial coordinates of the vectors ${\bf x}_i$ and ${\bf x}_j$, respectively, and $C$ and $t$ are the parameters of the measure. It can be seen that $s^{C,t}_\text{cfs}$ combines color similarity (first term) and spatial proximity (second term). The maximization of the aggregate combined fuzzy similarity in a particular neighborhood is equivalent to finding the vector that is  centrally located and is a good representative of the neighborhood in terms of chromatic content.
\par
The ordering criterion used in the directional-distance filter (DDF) \cite{Karakos97} combines the Minkowski and Cosine distances. In particular, the aggregate distance associated with pixel ${\bf x}_i$ is given by:
\begin{equation}
\label{equ_ddf}
D_i = \left( {\sum\limits_{j = 1}^n {d_p ({\bf x}_i ,{\bf x}_j )} } \right) \left( {\sum\limits_{j = 1}^n {d_\text{cosine} ({\bf x}_i ,{\bf x}_j )} } \right)
\end{equation}
Note that the idea of combining directional and magnitude processing also underlies the design of $s_\text{fmds}$. However, while $s_\text{fmds}$ is a proper similarity measure, i.e.\ it quantifies the similarity of two vectors, the ordering criterion given in \eqref{equ_ddf} is only defined in a particular neighborhood.

\begin{table}
\centering
\caption{ \label{tab_dist} Distance measures for R-ordering of color vectors ($\|{\bf x}\|$: Euclidean norm of ${\bf x}$) }
\begin{tabular}{ c|c }
\hline
\hline
Bray-Curtis Distance \cite{Bray57} & $d_\text{bray} ({\bf x}_i,{\bf x}_j) = \frac{{\sum\limits_{k = 1}^3 {\left| {x_{ik}  - x_{jk} } \right|} }}{{\sum\limits_{k = 1}^3 {x_{ik}  + x_{jk} } }}$\\
\hline
Canberra Distance \cite{Lance66} & $d_\text{canberra} ({\bf x}_i,{\bf x}_j) = \sum\limits_{k = 1}^3 {\frac{{\left| {x_{ik}  - x_{jk} } \right|}}{{x_{ik}  + x_{jk} }}}$\\
\hline
Chord Distance \cite{Borg05} & $d_\text{chord} ({\bf x}_i,{\bf x}_j) = \left( {\sum\limits_{k = 1}^3 {\left( {\sqrt {x_{ik} }  - \sqrt {x_{jk} } } \right)^2 } } \right)^{1/2}$\\ 
\hline
Cosine Distance \cite{Trahanias93} & $d_\text{cosine} ({\bf x}_i ,{\bf x}_j) = \theta  = \arccos \left( {\frac{{\sum\limits_{k = 1}^3 {x_{ik} x_{jk} } }}{{\left\| {{\bf x}_i } \right\|\left\| {{\bf x}_j } \right\|}}} \right)$\\
\hline
Divergence Coefficient \cite{Clark52} & $d_\text{divergence} ({\bf x}_i ,{\bf x}_j) = \left( {\sum\limits_{k = 1}^3 {\frac{{\left( {x_{ik}  - x_{jk} } \right)^2 }}{{x_{ik}  + x_{jk} }}} } \right)^{1/2}$\\
\hline
Goude Distance \cite{Goude72} & $d_\text{goude} ({\bf x}_i,{\bf x}_j) = \left( {\frac{{\left\| {{\bf x}_i } \right\|^2  + \left\| {{\bf x}_j } \right\|^2  - 2\sum\limits_{k = 1}^3 {x_{ik} x_{jk} } }}{{\left\| {{\bf x}_i } \right\|^2  + \left\| {{\bf x}_j } \right\|^2  + 2\sum\limits_{k = 1}^3 {x_{ik} x_{jk} } }}} \right)^{1/2}$\\
\hline
Soergel Distance \cite{Leach03} & $d_\text{soergel}({\bf x}_i,{\bf x}_j) = \frac{{\sum\limits_{k = 1}^3 {\left| {x_{ik} - x_{jk}} \right|} }}{{\sum\limits_{k = 1}^3 {\max (x_{ik} ,x_{jk})}}}$\\
\hline
Ware-Hedges Distance \cite{Ware78} & $d_\text{ware}({\bf x}_i,{\bf x}_j) = \sum\limits_{k = 1}^3 {\frac{{\left| {x_{ik} - x_{jk}} \right|}}{{\max (x_{ik},x_{jk})}}}$\\
\hline
\end{tabular}
\end{table}

A function $f: S \times S \rightarrow \mathbb{R}$ is called a metric on a set $S$ if it satisfies the following conditions for all $x,y,z \in S$ \cite{Korn00}:
\begin{enumerate}
  \item $f(x,y) \geq 0$ (non-negativity)
	\item $f(x,y)=0$ if and only if $x=y$ (self-identity)
	\item $f(x,y)=f(y,x)$ (symmetry)
	\item $f(x,y) \leq f(x,z) + f(y,z)$ (triangle inequality)
\end{enumerate}
Among the distance measures presented in this section, the Squared Euclidean, Bray-Curtis, and Goude measures are not metrics on $\mathbb{R}$ since they violate the triangle inequality. The Cosine distance is a semi (pseudo) metric on $\mathbb{R}$ because it satisfies all of the conditions except for the self-identity condition. On the other hand, it can be shown that \cite{Morillas05b,Morillas07a,Morillas07b} the fuzzy similarity measures $s_\text{fms}$, $s_\text{fds}$, $s_\text{fmds}$, and $s_\text{cfs}$ are stationary F-bounded fuzzy metrics in George and Veeramani's sense \cite{George94}.
\par
The operation counts for each distance measure are given in Table \ref{tab_cost} (ABS: absolute value, COMP: comparison, ADD: addition, SUB: subtraction, MULT: multiplication, DIV: division, SQRT: square root, ARCCOS: inverse cosine). Note that these operations are listed in ascending order according to their computational costs on a typical processor architecture, i.e.\ cost(ABS) $\leq$ cost(COMP) $\leq$ \ldots $\leq$ cost(SQRT) $\leq$ cost(ARCCOS). This may not hold on certain architectures, but it nevertheless permits cost comparisons between the measures to a certain extent. The actual cost of each operation is highly dependent on the implementation platform (the processor and the compiler).
\par
It should be noted that, even if it is mathematically simple, the use of a particular distance measure in an R-ordering based vector filter  can be computationally demanding. This is because during the filtering procedure the aggregate distance value \eqref{equ_agg_dist} is calculated for every vector ${\bf x}_i$ in each neighborhood $W(r,c)$. This means that the determination of the output vector ${\bf y}(r,c)$ requires ${{n(n - 1)} \mathord{\left/ {\vphantom {{n(n - 1)} 2}} \right. \kern-\nulldelimiterspace} 2}$ distance evaluations. For an $M \times N$ image this amounts to ${{MNn(n - 1)} \mathord{\left/ {\vphantom {{MNn(n - 1)} 2}} \right. \kern-\nulldelimiterspace} 2}$ evaluations. As an example, filtering a $512 \times 512$ image using a $3 \times 3$ window requires over $9.4$ million evaluations. Fortunately, the computation of certain distance measures can be accelerated using table lookups (see \S 
\ref{sec_exp}).

\begin{table}
\centering
\caption{ \label{tab_cost} Operation counts for the distance measures }
\begin{tabular}{ c|c|c|c|c|c|c|c }
\hline
Measure & ABS & COMP & ADD/SUB & MULT & DIV & SQRT & ARCCOS\\
\hline
\hline
$d_1$ & 3 & -- & 5 & -- & -- & -- & --\\
\hline
$d_2$ & -- & -- & 5 & 3 & -- & 1 & --\\
\hline
$d_2^2$ & -- & -- & 5 & 3 & -- & -- & --\\
\hline
$d_\infty$ & 3 & 2 & 3 & -- & -- & -- &\\
\hline
$d_\text{bray}$ & 3 & -- & 10 & -- & 1 & -- & --\\
\hline
$d_\text{canberra}$ & 3 & -- & 8 & -- & 3 & -- & --\\
\hline
$s_\text{cfs}$ & 2 & 1 & 9 & 4 & 2 & 1 & --\\
\hline
$d_\text{chord}$ & -- & -- & 5 & 3 & -- & 7 & --\\
\hline
$d_\text{cosine}$ & -- & -- & 6 & 10 & 1 & 1 & 1\\
\hline
$d_\text{divergence}$ & -- & -- & 8 & 3 & 3 & 1 & --\\
\hline
$s_\text{fds}$ & -- & 3 & 10 & 8 & 9 & 2 & --\\
\hline
$s_\text{fms}$ & -- & 3 & 6 & 2 & 3 & -- & --\\
\hline
$s_\text{fmds}$ & -- & 6 & 16 & 11 & 12 & 2 & --\\
\hline
$d_\text{goude}$ & -- & -- & 9 & 10 & 1 & 1 & --\\
\hline
$d_\text{soergel}$ & 3 & 3 & 7 & -- & 1 & -- & --\\
\hline
$d_\text{ware}$ & 3 & 3 & 5 & -- & 3 & -- & --\\
\hline
\end{tabular}
\end{table}

\section{Experimental Results and Discussion}
\label{sec_exp}

In order to evaluate the performance of the presented distance measures, a set of 100 high quality RGB images was collected from the Internet. The set included images of people, animals, plants, buildings, aerial maps, man-made objects, natural scenery, paintings, sketches, as well as scientific, biomedical, synthetic, and test images commonly used in the color image processing literature. The corruption in the images was simulated by the widely used correlated impulsive noise model \cite{Viero94}:
\begin{equation}
\label{eq_noise}
{\bf x} = \left\{ \begin{array}{l}
 {\bf o}\quad \quad \quad \qquad\, {\rm with}\,\,{\rm probability}\,\,1 - \varphi \,, \\ 
 \left\{ {r_1 ,o_2 ,o_3 } \right\}\quad {\rm with}\,\,{\rm probability}\,\,\varphi _1  \cdot \varphi , \\ 
 \left\{ {o_1 ,r_2 ,o_3 } \right\}\quad {\rm with}\,\,{\rm probability}\,\,\varphi _2  \cdot \varphi , \\ 
 \left\{ {o_1 ,o_2 ,r_3 } \right\}\quad {\rm with}\,\,{\rm probability}\,\,\varphi _3  \cdot \varphi , \\ 
 \left\{ {r_1 ,r_2 ,r_3 } \right\}\quad {\rm with}\,\,{\rm probability}\,\,\left( {1 - (\varphi _1  + \varphi _2  + \varphi _3 )} \right) \cdot \varphi  \\ 
 \end{array} \right.
\end{equation}
where ${\bf o} = \left\{ {o_1,o_2,o_3} \right\}$ and ${\bf x} = \left\{ {x_1,x_2,x_3} \right\}$ represent the original and noisy color vectors, respectively, ${\bf r} = \left\{ {r_1,r_2,r_3} \right\}$ is a random vector that represents the impulsive noise, $\varphi$ is the sample corruption probability, and $\varphi _1$, $\varphi _2$, and $\varphi _3$ are the corruption probabilities for the red, green, and blue channels, respectively. In the experiments, the channel corruption probabilities were set to 0.25.
\par
Filtering performance was evaluated by the following effectiveness criteria \cite{Celebi07}:
\begin{enumerate}
	\item Mean Absolute Error: $ \mbox{MAE} \left( {{\bf X},{\bf Y}} \right) = \frac{1}{{3MN}}\sum\limits_{r = 1}^M {\sum\limits_{c = 1}^N {d_1 \left( {{\bf x}(r,c),{\bf y}(r,c)} \right)} } $\\
	where ${\bf X}$ and ${\bf Y}$ denote respectively the $M \times N$ original and filtered images in the RGB color space. MAE measures the detail preservation capability of a filter.
	\item Mean Squared Error: $ \mbox{MSE} \left( {{\bf X},{\bf Y}} \right) = \frac{1}{{3MN}}\sum\limits_{r = 1}^M {\sum\limits_{c = 1}^N {d_2 ^2 \left( {{\bf x}(r,c),{\bf y}(r,c)} \right)} } $\\
	MSE measures the noise suppression capability of a filter.
	\item Normalized Color Difference: $ \mbox{NCD} \left( {{\bf X},{\bf Y}} \right) = \frac{{\sum\limits_{r = 1}^M {\sum\limits_{c = 1}^N {d_2 \left( {{\bf x}^\text{Lab} (r,c),{\bf y}^\text{Lab} (r,c)} \right)} } }}{{\sum\limits_{r = 1}^M {\sum\limits_{c = 1}^N {\left\| {{\bf x}^\text{Lab} (r,c)} \right\| } } }} $\\
	where ${\bf x}^\text{Lab}(r,c)$ and ${\bf y}^\text{Lab}(r,c)$ denote the CIEL*a*b* coordinates of the pixel $(r,c)$ in the original and filtered images, respectively. NCD measures the color preservation capability of a filter.
\end{enumerate}

Computational efficiency was measured by CPU time in milliseconds. All of the programs were implemented in the C language, compiled with the gcc 4.2.4 compiler, and executed on an Intel\textregistered Core\texttrademark2 Quad Q6700 2.66 GHz machine. The time measurements were  averaged over 10 identical runs. Although relative efficiency judgments can be made using the information given in Table \ref{tab_cost}, the actual execution times were measured for two main reasons. First, the cost of the basic operations is not uniform (for example, ARCCOS is computationally very expensive when compared to the others). Second, the use of lookup tables (LUTs) can significantly accelerate the computation of certain distance measures.
\par
Eighteen distance measures were evaluated in the experiments. These included the four variants of the Minkowski distance ($d_1$, $d_2$, $d_\infty$, and $d_2^2$), the eight measures listed in Table \ref{tab_dist}, and the four fuzzy similarity measures discussed in \S \ref{sec_dist}. Each of these measures was plugged into \eqref{equ_agg_dist} and the performance of the resulting filter was evaluated using the abovementioned criteria. A square window of size $3 \times 3$ ($n = 9$) is used in all of the filters. Although the ordering criterion used in DDF \eqref{equ_ddf} is not a distance measure, because of its widespread use in the literature, it was included in the experiments. Furthermore, the identity filter (indicated by the label 'none'), which performs no filtering was included as a baseline. The $K$ parameters of $s_\text{fms}$ and $s_\text{fds}$ were set to 1024 and 4 \cite{Morillas07b}, respectively. The $C$ and $t$ parameters of $s_\text{cfs}$ were set to $150$ and $4$ \cite{Morillas07a}, respectively.
\par
Implementation details for the distance measures are given below ($\vec{0} = [0,0,0]$):

\begin{itemize}
\renewcommand{\labelitemi}{$\triangleright$}
  \item $d_2^2$: Instead of calculating ${{n(n - 1)} \mathord{\left/ {\vphantom {{n(n - 1)} 2}} \right. \kern-\nulldelimiterspace} 2}$ distance values in each neighborhood $W(r,c)$, first the mean vector $\bar {\bf x} = {1 \mathord{\left/ {\vphantom {1 n}} \right. \kern-\nulldelimiterspace} n}\sum\nolimits_{i = 1}^n {{\bf x}_i}$ is calculated and then the output vector is determined to be the input vector that has the minimum $d_2^2$ distance to the mean vector, i.e.\ ${\bf y}(r,c) = \mathop {{\rm argmin}}\limits_{{\bf x}_i \in W(r,c)} d_2^2({\bf x}_i, \bar {\bf x})$ \cite{Barnett76}.
	\item $d_\text{bray}$, $d_\text{goude}$, $d_\text{soergel}$: If ${\bf x}_i = {\bf x}_j = \vec{0}$, the distance between the two vectors is taken as $0$.
	\item $d_\text{canberra}$, $d_\text{divergence}$, $d_\text{ware}$: If $x_{ik} = x_{jk} = 0$ for any component $k \in \{1,2,3\}$, the contribution of that component to the distance between ${\bf x}_i$ and ${\bf x}_j$ is ignored. The value of the term in the summation is precomputed and stored in a 2D LUT of size $256 \times 256$.
	\item $s_\text{cfs}$: The spatial proximity term is precomputed and stored in a 2D LUT of size $n \times n$.
	\item $d_\text{chord}$: The value of each term in the summation is precomputed and stored in a 1D LUT of size $256$.
	\item $d_\text{cosine}$: If either ${\bf x}_i = \vec{0}$ or ${\bf x}_j = \vec{0}$, the distance between the two vectors is taken as $\pi$ (maximum possible angular distance). If ${\bf x}_i = {\bf x}_j = \vec{0}$, the distance between them is taken as $0$ (minimum possible angular distance).
	\item $s_\text{fds}$: If either ${\bf x}_i = \vec{0}$ or ${\bf x}_j = \vec{0}$, the similarity between the two vectors is taken as $0$ (minimum possible similarity). If ${\bf x}_i = {\bf x}_j = \vec{0}$, the similarity between them is taken as $1$ (maximum possible similarity).
	\item $s_\text{fms}$: The value of the term in the product is precomputed and stored in a 2D LUT of size $256 \times 256$.
\end{itemize}

Table \ref{tab_quality} shows the average effectiveness ranking of the distance measures over the entire image set. Note that the ranks start from $0$ and the smaller the rank value, the better the corresponding distance measure. For example, with respect to the MAE criterion, the $s_\text{fmds}$ measure has an average rank of $2.00$ at 20\% noise level. In other words, with respect to detail preservation, the R-ordering based vector filter that utilizes $s_\text{fmds}$ ranks, on the average, in top 3 among 18 filters. The last column shows the overall mean effectiveness rank for each distance measure. It can be seen that, on the average, $s_\text{cfs}$, $s_\text{fmds}$, and $s_\text{fms}$ perform the best. Interestingly, these are all fuzzy metrics, which indicates that fuzzy logic might be better suited for outlier detection when compared to traditional distance measures. Among these metrics, $s_\text{cfs}$ outperforms the other two by a large margin as evidenced by its impressive $0.33$ average rank. The success of $s_\text{cfs}$ is most likely due to its consideration of spatial proximity, an idea that seems to be overlooked in the literature. It should also be noted that the top two filters are both hybrid in nature: $s_\text{cfs}$ combines magnitude similarity with spatial proximity, whereas $s_\text{fmds}$ combines magnitude similarity with directional similarity. In addition to the fuzzy metrics, $d_1$ and $d_\text{divergence}$ also perform better than $d_2$, which is the most commonly used distance measure in the literature \cite{Smolka06,Plataniotis00,Smolka04,Lukac05,Lukac06,Celebi07}. Despite the fact that it differs from $d_2$ only by its lack of a square root operation, $d_2^2$ performs among the worst. This is not surprising given the fact that the R-ordering based vector filter that utilizes $d_2^2$ essentially approximates the average filter \cite{Barni92}, which is known to blur the fine details. $d_\text{cosine}$ and $s_\text{fds}$ also perform poorly, which shows that directional information alone is not sufficient to determine the most representative vector in a neighborhood. Interestingly, the identity filter ('none'), which leaves the noisy image unchanged preserves the fine details better than $s_\text{fds}$, $d_2^2$, and $d_\text{cosine}$ at 10\% noise level.
\par
It should be noted that the performance of the fuzzy metrics can be enhanced by tuning their parameters adaptively according to the image characteristics and the level of noise, see, for example, \cite{Morillas07a}.
\par
\begin{figure}[!ht]
\centering
 \subfigure[original]{\label{peppers_a}\includegraphics[width=0.28\columnwidth]{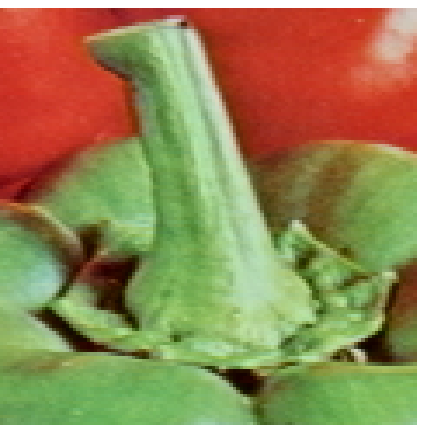}}
 \hspace{0.16in}
 \subfigure[10\% noisy (MAE:6.34, MSE:1037.19)]{\label{peppers_b}\includegraphics[width=0.28\columnwidth]{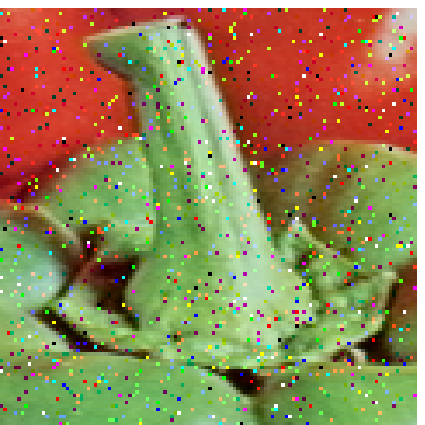}}
 \hspace{0.16in}
 \subfigure[$d_2^2$ (MAE:3.72, MSE:42.88)]{\label{peppers_c}\includegraphics[width=0.28\columnwidth]{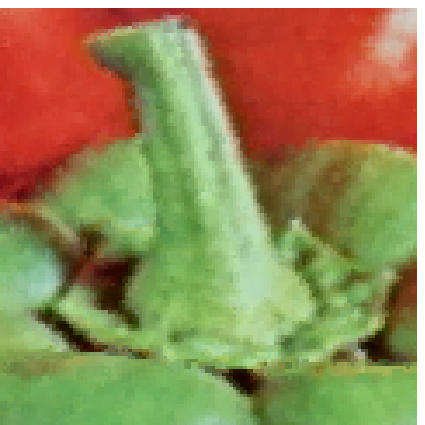}}
 \\
 \subfigure[$d_\text{canberra}$ (MAE:2.35, MSE:26.36)]{\label{peppers_d}\includegraphics[width=0.28\columnwidth]{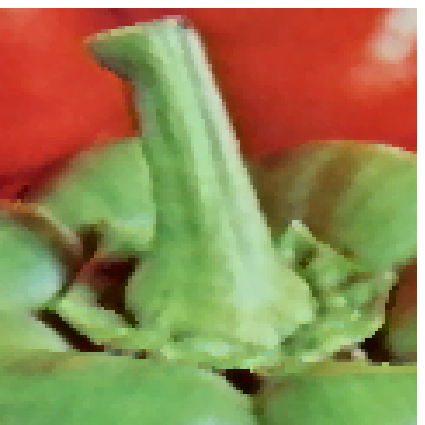}}
 \hspace{0.16in}
 \subfigure[$d_\text{divergence}$ (MAE:2.29, MSE:20.71)]{\label{peppers_e}\includegraphics[width=0.28\columnwidth]{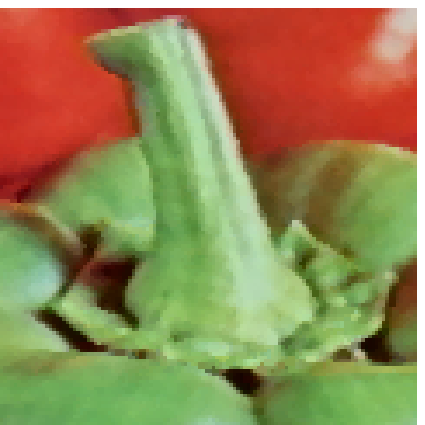}}
 \hspace{0.16in}
 \subfigure[$d_2$ (MAE:2.29, MSE:20.77)]{\label{peppers_f}\includegraphics[width=0.28\columnwidth]{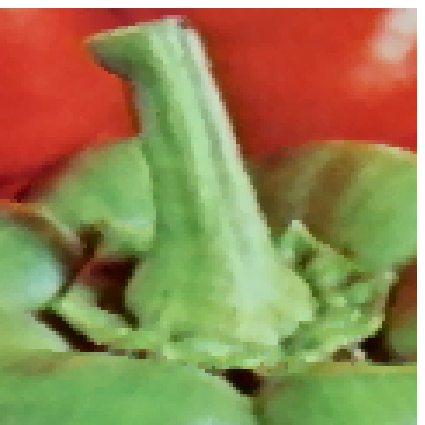}}
 \\
 \subfigure[$d_1$ (MAE:2.23, MSE:20.70)]{\label{peppers_g}\includegraphics[width=0.28\columnwidth]{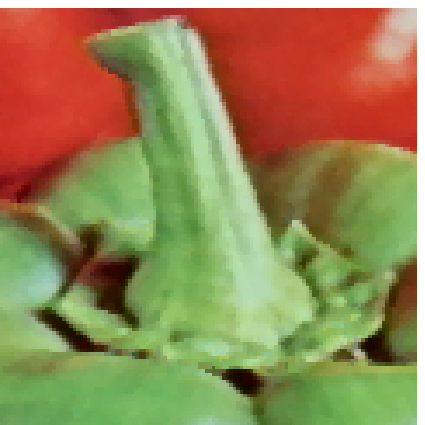}}
 \hspace{0.16in}
 \subfigure[$s_\text{fms}$ (MAE:2.22, MSE:20.42)]{\label{peppers_h}\includegraphics[width=0.28\columnwidth]{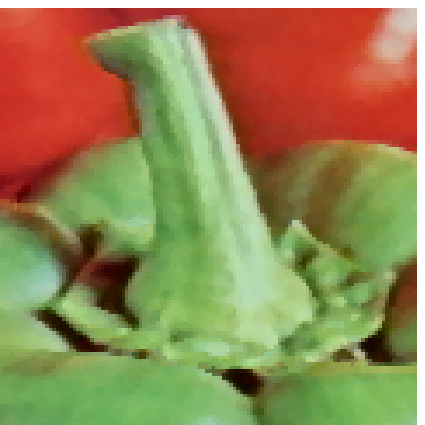}}
 \hspace{0.16in}
 \subfigure[$s_\text{cfs}$ (MAE:0.49, MSE:11.06)]{\label{peppers_i}\includegraphics[width=0.28\columnwidth]{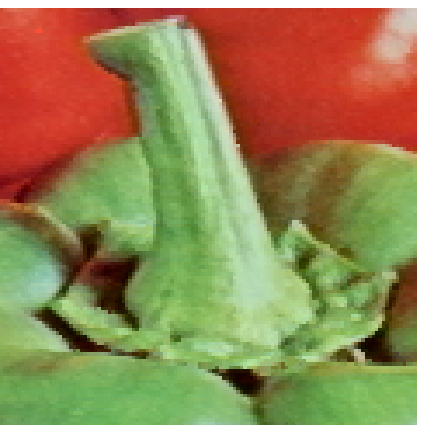}}
 \caption{Filtering results for the Peppers image corrupted with 10\% noise}
 \label{fig_peppers}
\end{figure}

Figure \ref{fig_peppers} shows sample filtering results for a close-up part of the Peppers image contaminated with 10\% noise. It can be seen that $d_2^2$ blurs the fine details severely. $d_\text{divergence}$, $d_2$, $d_1$, and $s_\text{fms}$ are fairly close in terms of detail preservation and noise removal performance. On the other hand, $s_\text{cfs}$ performs significantly better than the others, producing an output image that closely resembles the original image.
\par
\begin{figure}[!ht]
\centering
 \subfigure[original]{\label{lenna_a}\includegraphics[width=0.28\columnwidth]{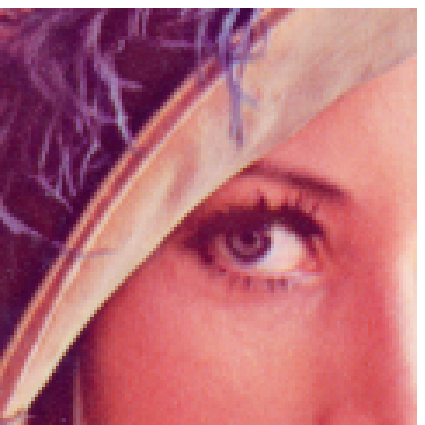}}
 \hspace{0.16in}
 \subfigure[20\% noisy (MAE:12.67, MSE:1960.48)]{\label{lenna_b}\includegraphics[width=0.28\columnwidth]{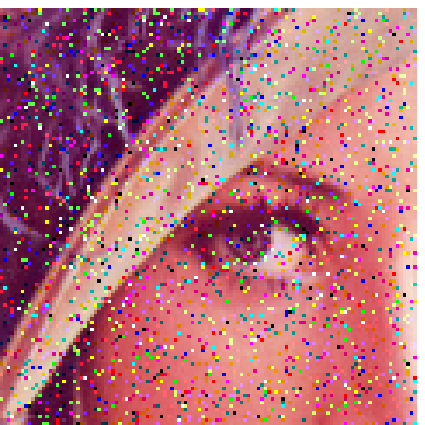}}
 \hspace{0.16in}
 \subfigure[$d_\text{cosine}$ (MAE:4.23, MSE:55.06)]{\label{lenna_c}\includegraphics[width=0.28\columnwidth]{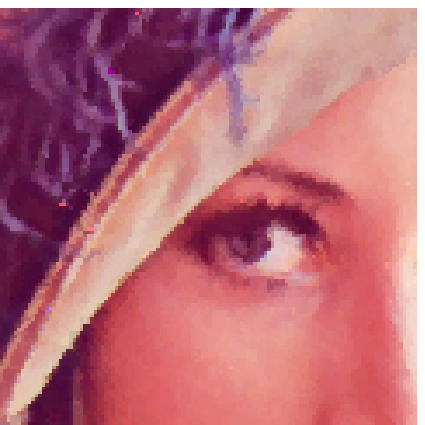}}
 \\
 \subfigure[$d_\infty$ (MAE:4.01, MSE:48.72)]{\label{lenna_d}\includegraphics[width=0.28\columnwidth]{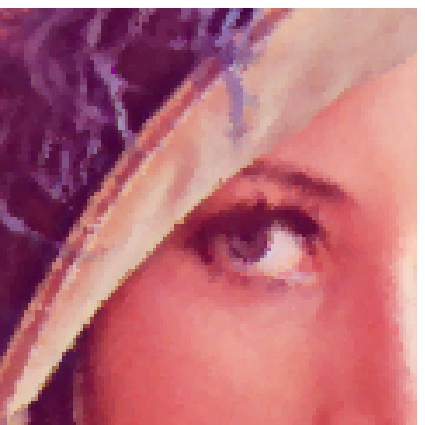}}
 \hspace{0.16in}
 \subfigure[$d_\text{goude}$ (MAE:3.75, MSE:42.51)]{\label{lenna_e}\includegraphics[width=0.28\columnwidth]{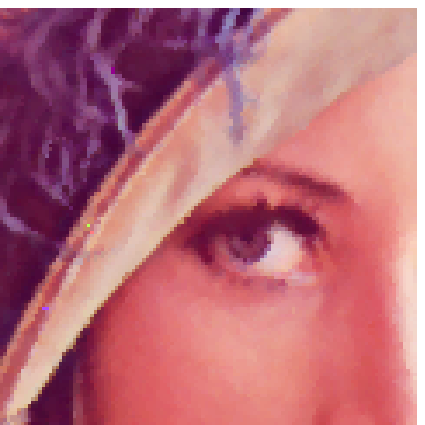}}
 \hspace{0.16in}
 \subfigure[$d_2$ (MAE:3.70, MSE:40.71)]{\label{lenna_f}\includegraphics[width=0.28\columnwidth]{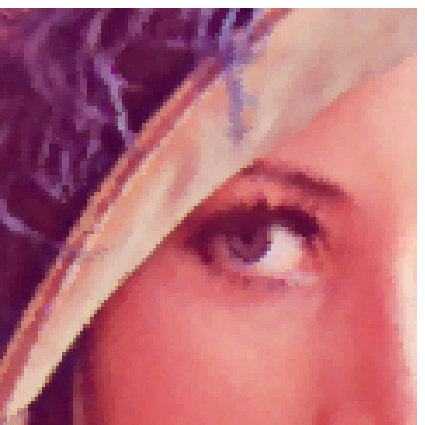}}
 \\
 \subfigure[$s_\text{fmds}$ (MAE:3.54, MSE:37.48)]{\label{lenna_g}\includegraphics[width=0.28\columnwidth]{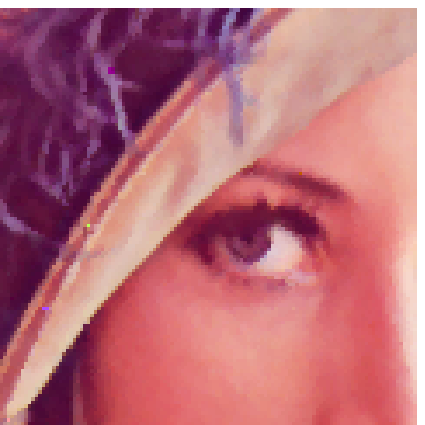}}
 \hspace{0.16in}
 \subfigure[$d_1$ (MAE:3.49, MSE:37.72)]{\label{lenna_h}\includegraphics[width=0.28\columnwidth]{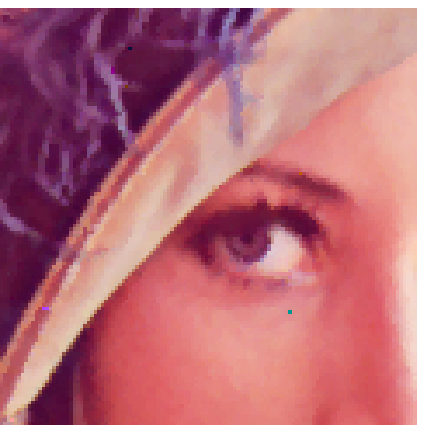}}
 \hspace{0.16in}
 \subfigure[$s_\text{cfs}$ (MAE:1.26, MSE:23.24)]{\label{lenna_i}\includegraphics[width=0.28\columnwidth]{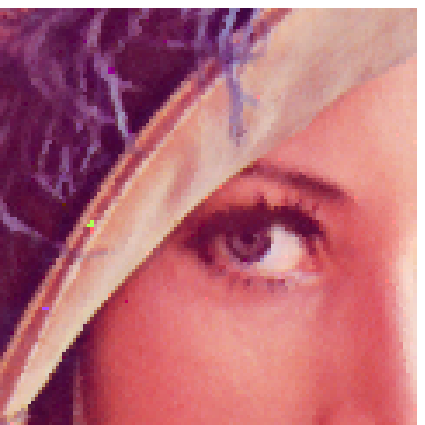}}
 \caption{Filtering results for the Lenna image corrupted with 20\% noise}
 \label{fig_lenna}
\end{figure}

Figure \ref{fig_lenna} shows sample filtering results for a close-up part of the Lenna image contaminated with 20\% noise. It can be seen that $d_\text{cosine}$ not only fails to remove a significant amount of noise, but also smears the fine details. Even though $d_\text{goude}$, $d_2$, $s_\text{fmds}$, and $d_1$ suppress the noise well, this comes at the expense of the blurring of the fine details, e.g. the boa fur and the eye lashes. In contrast, $s_\text{cfs}$ achieves a remarkable balance between detail preservation and noise removal. 
\par
\begin{figure}[!ht]
\centering
 \subfigure[original]{\label{baboon_a}\includegraphics[width=0.28\columnwidth]{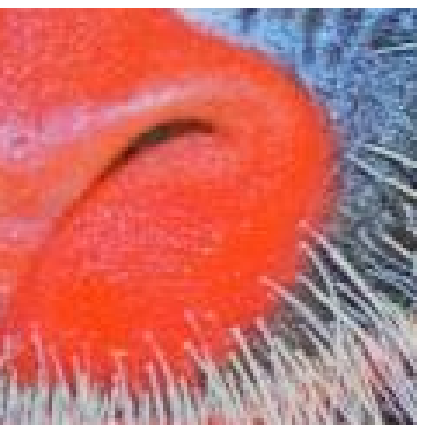}}
 \hspace{0.16in}
 \subfigure[30\% noisy (MAE:19.07, MSE:2894.43)]{\label{baboon_b}\includegraphics[width=0.28\columnwidth]{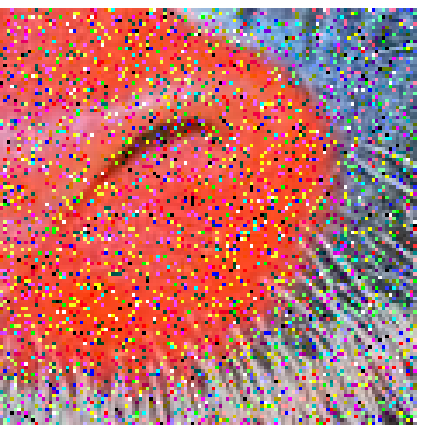}}
 \hspace{0.16in}
 \subfigure[$d_\text{chord}$ (MAE:11.86, MSE:383.68)]{\label{baboon_c}\includegraphics[width=0.28\columnwidth]{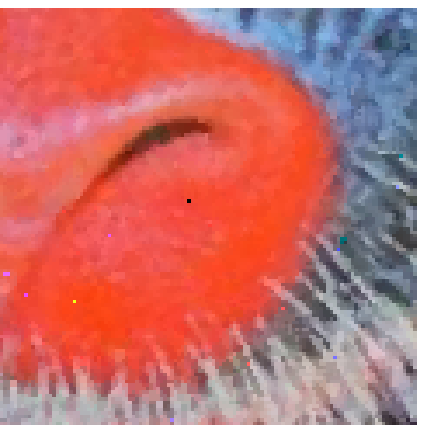}}
 \\
 \subfigure[$d_2$ (MAE:11.76, MSE:375.17)]{\label{baboon_d}\includegraphics[width=0.28\columnwidth]{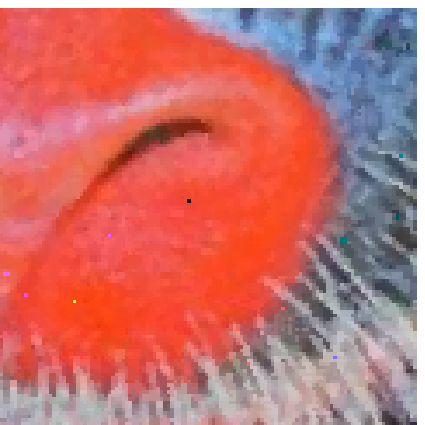}}
 \hspace{0.16in}
 \subfigure[$d_\text{ware}$ (MAE:11.66, MSE:436.98)]{\label{baboon_e}\includegraphics[width=0.28\columnwidth]{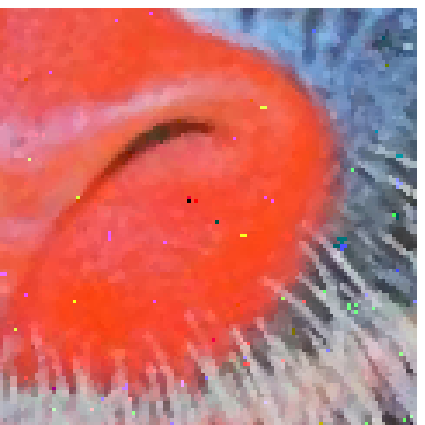}}
 \hspace{0.16in}
 \subfigure[$d_\text{soergel}$ (MAE:11.61, MSE:410.25)]{\label{baboon_f}\includegraphics[width=0.28\columnwidth]{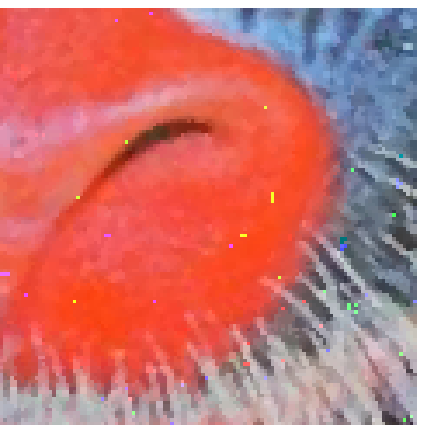}}
 \\
 \subfigure[$d_1$ (MAE:11.29, MSE:363.30)]{\label{baboon_g}\includegraphics[width=0.28\columnwidth]{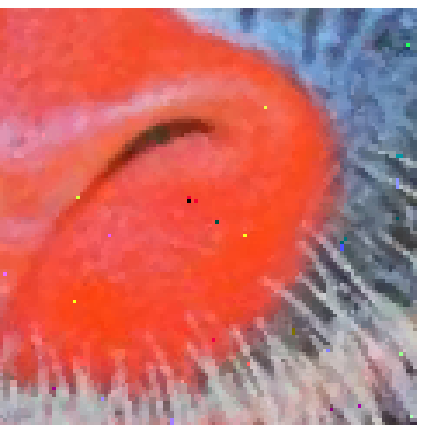}}
 \hspace{0.16in}
 \subfigure[$s_\text{fms}$ (MAE:11.22, MSE:361.53)]{\label{baboon_h}\includegraphics[width=0.28\columnwidth]{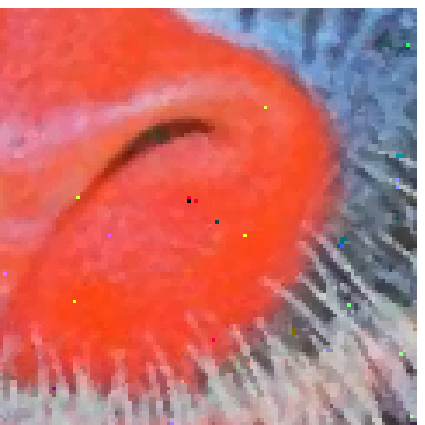}}
 \hspace{0.16in}
 \subfigure[$s_\text{cfs}$ (MAE:7.72, MSE:337.35)]{\label{baboon_i}\includegraphics[width=0.28\columnwidth]{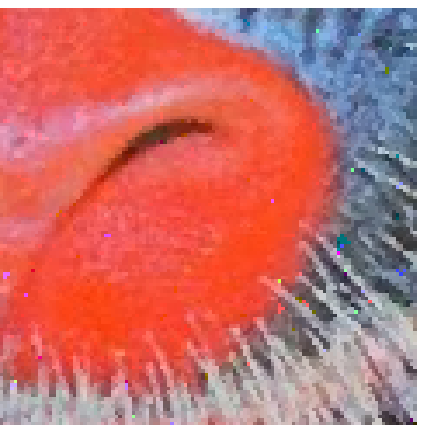}}
 \caption{Filtering results for the Baboon image corrupted with 30\% noise}
 \label{fig_baboon}
\end{figure}

Figure \ref{fig_baboon} shows sample filtering results for a close-up part of the Baboon image contaminated with 30\% noise. This image  presents a challenging case not only because of its high level of noise, but also due to its complex structural content, e.g.\ the whiskers. It can be seen that $d_\text{chord}$ and $d_2$ fail to preserve the fine details, producing output images with blurry and/or broken whiskers. $d_\text{ware}$, $d_\text{soergel}$, $d_1$, and $s_\text{fms}$ preserve the details better, while leaving many noisy pixels intact. As before, $s_\text{cfs}$ outperforms the others in terms of both detail preservation and noise removal.
\par
Table \ref{tab_time} shows the average efficiency ranking of the distance measures. Obviously, the identity filter consistently ranks first since it involves no actual filtering. As expected, $d_2^2$ is the most efficient distance measure since, in contrast to the other measures, it does not involve ${{n(n - 1)} \mathord{\left/ {\vphantom {{n(n - 1)} 2}} \right. \kern-\nulldelimiterspace} 2}$ distance evaluations in each neighborhood. $d_1$ is the second most efficient measure as it involves only the simplest mathematical operations, i.e.\ absolute value and addition/subtraction. The computation of $d_1$ can be further accelerated using the method described in \cite{Barni97}. $d_\text{ware}$, $s_\text{fms}$, and $d_\text{canberra}$ are also among the most efficient since they can be calculated rapidly using LUTs. Except for $s_\text{fms}$, the fuzzy metrics are not particularly efficient. Among these, $s_\text{cfs}$ and $s_\text{fmds}$ partially benefit from the use of LUTs. However, the calculation of the color similarity term in $s_\text{cfs}$ and the directional similarity term in $s_\text{fmds}$ increases the computational requirements for these filters. Finally, $d_\text{cosine}$ and {\small ddf} are the least efficient of all, which is due to their use of the computationally expensive ARCCOS operation.
\par
Table \ref{tab_time_peppers} shows the average computational time requirements of the distance measures on the $512 \times 512$ Peppers image over 100 identical runs. The last two columns give the actual (measured in seconds) and relative (indicated in units of {\scshape t}) computational times for each distance measure. It can be seen that the actual time differences among many of the measures are rather negligible. However, $d_\text{cosine}$ and {\small ddf} are still relatively time consuming when compared to the others.

\begin{table} 
\centering
\caption{ \label{tab_quality} Average effectiveness ranking of the distance measures }
\begin{tabular}{ c|ccc|ccc|ccc|c }
\hline
\multirow{2}{*}{Measure} & \multicolumn{3}{|c|}{MAE} & \multicolumn{3}{|c|}{MSE} & \multicolumn{3}{|c|}{NCD} & \multirow{2}{*}{Mean}\\
 & 10\% & 20\% & 30\% & 10\% & 20\% & 30\% & 10\% & 20\% & 30\% & \\
\hline
\hline
$s_\text{cfs}$ & 0.03 & 0.00 & 0.00 & 0.00 & 0.25 & 2.67 & 0.00 & 0.00 & 0.04 & 0.33\\
\hline
$s_\text{fmds}$ & 2.44 & 2.00 & 2.23 & 3.42 & 2.42 & 2.01 & 1.47 & 1.21 & 1.30 & 2.06\\
\hline
$s_\text{fms}$ & 2.20 & 1.65 & 1.30 & 2.07 & 1.79 & 1.62 & 6.49 & 3.21 & 2.53 & 2.54\\
\hline
$d_1$ & 3.45 & 2.86 & 2.63 & 1.99 & 2.42 & 2.61 & 8.13 & 4.90 & 4.06 & 3.67\\
\hline
$d_\text{divergence}$ & 6.23 & 6.08 & 4.85 & 4.67 & 3.74 & 2.94 & 3.38 & 3.25 & 3.21 & 4.26\\
\hline
$d_2$ & 7.61 & 7.98 & 7.24 & 5.97 & 5.01 & 4.18 & 4.12 & 5.46 & 5.34 & 5.88\\
\hline
$d_\text{chord}$ & 9.78 & 9.69 & 8.78 & 7.86 & 6.58 & 5.53 & 6.92 & 7.66 & 6.76 & 7.73\\
\hline
$d_\text{canberra}$ & 6.10 & 5.24 & 6.03 & 6.48 & 8.35 & 10.43 & 10.32 & 8.97 & 10.46 & 8.04\\
\hline
{\small ddf} & 9.53 & 9.99 & 9.13 & 11.00 & 9.05 & 7.28 & 4.95 & 5.85 & 5.84 & 8.07\\
\hline
$d_\text{bray}$ & 6.03 & 6.40 & 8.26 & 6.32 & 9.19 & 11.24 & 10.16 & 10.05 & 11.14 & 8.75\\
\hline
$d_\text{goude}$ & 9.30 & 11.08 & 11.61 & 8.66 & 9.43 & 8.94 & 5.84 & 8.85 & 8.99 & 9.19\\
\hline
$d_\text{soergel}$ & 9.02 & 7.64 & 7.21 & 10.25 & 9.82 & 10.46 & 10.90 & 8.89 & 8.80 & 9.22\\
\hline
$d_\infty$ & 12.19 & 12.59 & 12.53 & 11.11 & 11.16 & 9.25 & 6.15 & 10.81 & 10.01 & 10.64\\
\hline
$d_\text{ware}$ & 10.49 & 8.35 & 9.39 & 11.67 & 12.36 & 13.51 & 12.95 & 12.24 & 13.12 & 11.56\\
\hline
$s_\text{fds}$ & 14.75 & 14.12 & 14.09 & 14.90 & 14.73 & 14.72 & 13.93 & 13.81 & 13.65 & 14.30\\
\hline
$d_2^2$ & 15.90 & 15.73 & 15.64 & 14.11 & 14.03 & 12.76 & 15.83 & 15.88 & 15.80 & 15.08\\
\hline
$d_\text{cosine}$ & 15.45 & 15.12 & 15.16 & 15.52 & 15.67 & 15.85 & 14.73 & 14.96 & 14.95 & 15.27\\
\hline
{\small none} & 12.50 & 16.48 & 16.92 & 17.00 & 17.00 & 17.00 & 16.73 & 17.00 & 17.00 & 16.40\\
\hline
\end{tabular}
\end{table}

\begin{table} 
\centering
\caption{ \label{tab_time} Average efficiency ranking of the distance measures }
\begin{tabular}{ c|ccc|c }
\hline
\multirow{2}{*}{Measure} & \multicolumn{3}{|c|}{Time} & \multirow{2}{*}{Mean}\\
 & 10\% & 20\% & 30\% & \\
\hline
\hline
{\small none} & 0.00 & 0.00 & 0.00 & 0.00\\
\hline
$d_2^2$ & 1.00 & 1.00 & 1.00 & 1.00\\
\hline
$d_1$ & 2.00 & 2.00 & 2.00 & 2.00\\
\hline
$d_\text{ware}$ & 3.48 & 3.42 & 3.47 & 3.46\\
\hline
$s_\text{fms}$ & 3.52 & 3.58 & 3.54 & 3.55\\
\hline
$d_\text{canberra}$ & 5.00 & 5.00 & 4.99 & 5.00\\
\hline
$d_\infty$ & 6.00 & 6.01 & 6.00 & 6.00\\
\hline
$d_\text{bray}$ & 7.00 & 6.99 & 7.00 & 7.00\\
\hline
$d_\text{chord}$ & 8.61 & 8.30 & 8.09 & 8.33\\
\hline
$d_\text{soergel}$ & 8.41 & 8.71 & 8.92 & 8.68\\
\hline
$d_2$ & 10.02 & 10.01 & 10.08 & 10.04\\
\hline
$d_\text{divergence}$ & 10.98 & 11.00 & 10.93 & 10.97\\
\hline
$s_\text{cfs}$ & 12.06 & 12.05 & 12.06 & 12.06\\
\hline
$d_\text{goude}$ & 12.96 & 12.97 & 12.96 & 12.96\\
\hline
$s_\text{fds}$ & 13.97 & 13.97 & 13.97 & 13.97\\
\hline
$s_\text{fmds}$ & 14.99 & 14.99 & 14.99 & 14.99\\
\hline
$d_\text{cosine}$ & 16.00 & 16.00 & 16.00 & 16.00\\
\hline
{\small ddf} & 17.00 & 17.00 & 17.00 & 17.00\\
\hline
\end{tabular}
\end{table}

\begin{table} 
\centering
\caption{ \label{tab_time_peppers} Time requirements of the distance measures on the Peppers image }
\begin{tabular}{ c|c|c }
\hline
Measure & Actual Time & Relative Time\\
\hline
\hline
{\small none} & 0.000 & 0{\scshape t}\\
\hline
$d_2^2$ & 0.063 & 1{\scshape t}\\
\hline
$d_1$ & 0.153 & 2{\scshape t}\\
\hline
$s_\text{fms}$ & 0.181 & 3{\scshape t}\\
\hline
$d_\text{ware}$ & 0.182 & 3{\scshape t}\\
\hline
$d_\text{canberra}$ & 0.183 & 3{\scshape t}\\
\hline
$d_\infty$ & 0.240 & 4{\scshape t}\\
\hline
$d_\text{bray}$ & 0.256 & 4{\scshape t}\\
\hline
$d_\text{soergel}$ & 0.346 & 6{\scshape t}\\
\hline
$d_\text{chord}$ & 0.356 & 6{\scshape t}\\
\hline
$d_2$ & 0.366 & 6{\scshape t}\\
\hline
$d_\text{divergence}$ & 0.396 & 6{\scshape t}\\
\hline
$s_\text{cfs}$ & 0.508 & 8{\scshape t}\\
\hline
$d_\text{goude}$ & 0.533 & 8{\scshape t}\\
\hline
$s_\text{fds}$ & 0.623 & 10{\scshape t}\\
\hline
$s_\text{fmds}$ & 0.694 & 11{\scshape t}\\
\hline
$d_\text{cosine}$ & 1.424 & 23{\scshape t}\\
\hline
{\small ddf} & 1.684 & 27{\scshape t}\\
\hline
\end{tabular}
\end{table}

\section{Conclusions and Future Work}
\label{sec_conc}
In this paper, we reviewed various distance measures that can be used to order color vectors in R-ordering based vector filters. We evaluated the performance of these measures on a large and diverse set of images using several effectiveness and efficiency criteria. The results demonstrated that the fuzzy similarity metrics $s_\text{cfs}$, $s_\text{fmds}$, and $s_\text{fms}$ significantly outperform the commonly used $d_1$ and $d_2$ metrics in terms of detail preservation, noise removal, and color preservation. Furthermore, these fuzzy metrics are parametrized, which permits the tuning of their parameters according to the image characteristics and the level of noise.
\par
The use of the presented distance measures is not limited to the basic R-ordering based vector filters such as VMF, BVDF, and DDF. Each of these basic filters can in turn be used as a back-end in the implementation of more advanced switching filters, i.e.\ those filters that selectively remove the noise by utilizing impulse detectors \cite{Smolka06,Lukac06,Celebi07}. However, a distance measure that performs well in a non-switching noise removal filter may not necessarily perform as well when combined with an (imperfect) impulse detector. In other words, the complex interaction between the front-end (impulse detector) and back-end (noise removal filter) in a switching filter will likely to have a non-negligible influence on the ultimate performance of a distance measure. This issue will be investigated in a future study.

\section*{Acknowledgments}
This publication was made possible by a grant from The Louisiana Board of Regents (LEQSF2008-11-RD-A-12). The author is grateful to the anonymous reviewers for their insightful suggestions and constructive comments that improved the quality and presentation of this paper.

\bibliographystyle{IEEEbib}
\bibliography{vmf_metrics_bib}

\end{document}